\documentclass[10pt,twocolumn,letterpaper]{article}

\usepackage{cvpr}              %

\usepackage{graphicx}
\usepackage{amsmath}
\usepackage{amssymb}
\usepackage{booktabs}
\usepackage{multirow}
\usepackage{tabularx}
\usepackage{makecell}
\usepackage[table]{xcolor}
\usepackage[linesnumbered,ruled,vlined]{algorithm2e}
\usepackage[accsupp]{axessibility} %

\newcommand{\mycomment}[1]{}
\usepackage[pagebackref,breaklinks,colorlinks]{hyperref}

\usepackage[capitalize]{cleveref}
\crefname{section}{Sec.}{Secs.}
\Crefname{section}{Section}{Sections}
\Crefname{table}{Table}{Tables}
\crefname{table}{Tab.}{Tabs.}

\def\cvprPaperID{17} %
\def\confName{CVPR}
\def\confYear{2024}

\begin{document}

\title{Can the accuracy bias by facial hairstyle be reduced \\ through balancing the training data?}

\author{Kagan Ozturk, Haiyu Wu, Kevin W. Bowyer\\
University of Notre Dame\\
\tt\small \{kztrk, hwu6, kwb\}@nd.edu
}

\maketitle

\begin{abstract}
Appearance of a face can be greatly altered by growing a beard and mustache.  The facial hairstyles in a pair of images can cause marked changes to the impostor distribution and the genuine distribution. Also, different distributions of facial hairstyle across demographics could cause a false impression of  relative accuracy across demographics. We first show that, even though larger training sets boost the recognition accuracy on all facial hairstyles, accuracy variations caused by facial hairstyles persist regardless of the size of the training set.  Then, we analyze the impact of having different fractions of the training data represent facial hairstyles. We created balanced training sets using a set of identities available in Webface42M that both have clean-shaven and facial hair images. We find that, even when a face recognition model is trained with a balanced clean-shaven / facial hair training set, accuracy variation on the test data does not diminish.  Next, data augmentation is employed to further investigate the effect of facial hair distribution in training data by manipulating facial hair pixels with the help of facial landmark points and a facial hair segmentation model. Our results show facial hair causes an accuracy gap between clean-shaven and facial hair images, and this impact can be significantly different between African-Americans and Caucasians. 
\end{abstract}

\section{Introduction}
\label{sec:intro}

While deep CNN face matchers have achieved great success \cite{schroff2015facenet, deng2019arcface, kim2022adaface, meng2021magface, parkhi2015deep, taigman2014deepface, wang2018cosface}, researchers are also investigating concerns about fairness or bias in accuracy \cite{hupont2019demogpairs, robinson2020face, drozdowski2020demographic, xu2021consistent, wu2023should, jain2021biometrics, robinson2023balancing, grother2019face}. Because deep CNN face matchers are trained with huge web-scraped datasets and the decision-making of the final model is not transparent, the imbalance of demographic groups in the training data is often the ``knee-jerk’’ first suspect for the cause of any bias.

However, it has been shown that having a training set that is explicitly balanced on the number of subjects and images is not sufficient to mitigate, for example, the accuracy difference across gender \cite{albiero2020does}. The correlation of various facial attributes with recognition accuracy differences has also been extensively investigated \cite{terhorst2021comprehensive}. Moreover, bias measurement methods are analyzed in recent works \cite{serna2021insidebias, cavazos2020accuracy, fabbrizzi2022survey}.
\begin{figure}[t]
            \includegraphics[width=.98\linewidth]{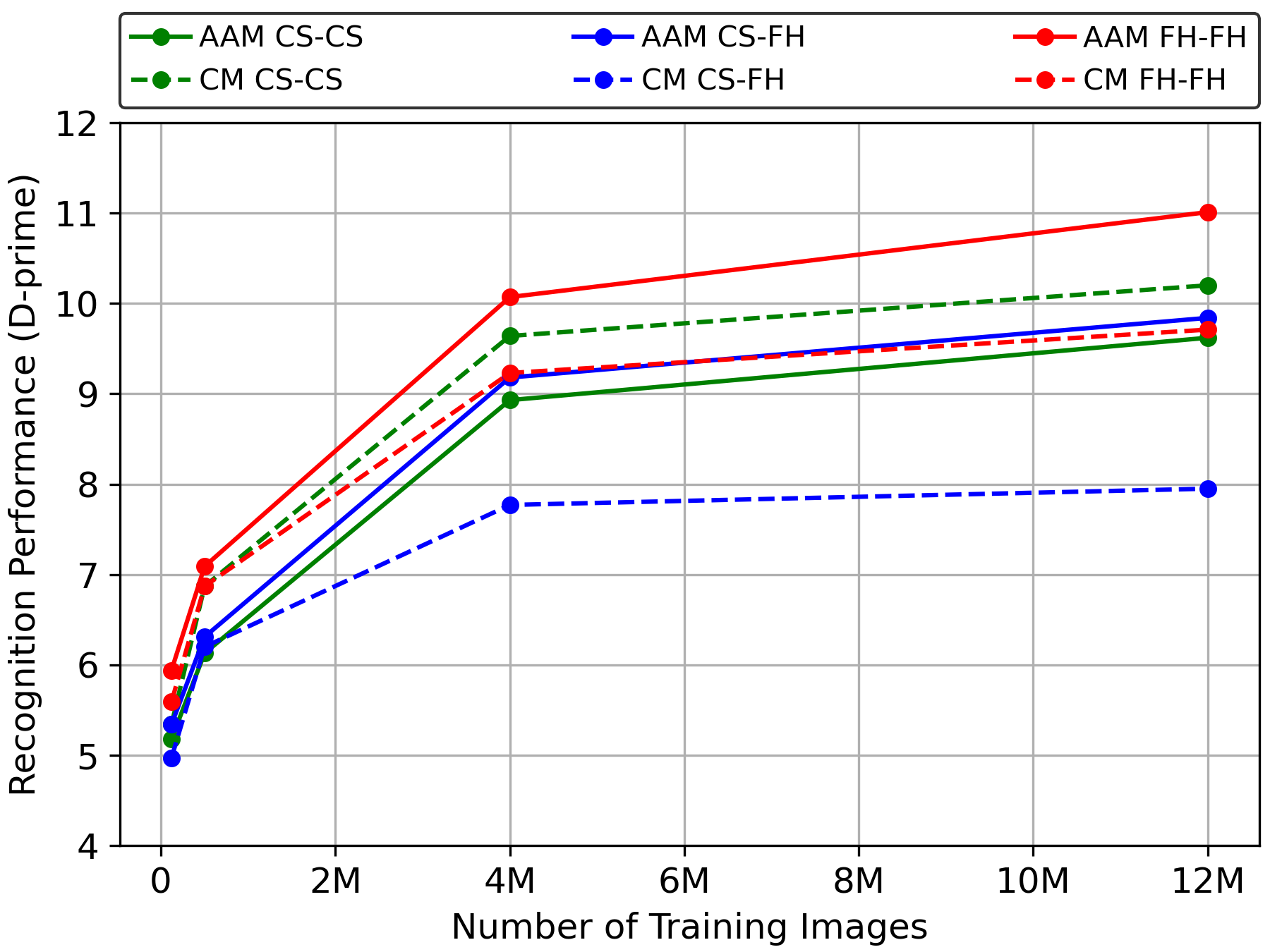}
    \caption{The effect of training size on recognition is given for African-American (AAM) and Caucasian (CM) males on MORPH using  4 training sets: a facial hair-balanced subset of WebFace42M (120K images Sec. \ref{sec:within subjects}), Casia-WebFace (500K images), WebFace4M (4M images) and WebFace12M (12M images). Higher d-prime between genuine and impostor distribution means better recognition accuracy. While d-prime values for CS-CS (clean-shaven v. clean-shaven), CS-FH (clean-shaven v. facial hair) and FH-FH (facial hair v. facial hair) image pairs   consistently increase as training data gets larger, 
    the d-prime gap across facial hairstyles also increases. AdaFace \cite{kim2022adaface} loss is used to train the models.}
    \label{fig:training_size}

\end{figure}

Among the facial attributes, facial hairstyle (e.g., beard and mustache) can be easily changed to alter one’s appearance.  Facial hairstyle choices are influenced by a myriad of factors, including cultural norms, genetic factors and fashion trends.  Facial hairstyle can occlude a substantial portion of the face and thereby affect the feature vector and similarity score produced by a face recognition model.  This work seeks to understand how the facial hairstyles present in a pair of images affect the impostor and genuine distribution, and whether the training data can be engineered to minimize accuracy differences across facial hairstyles.  While the effect of facial hair on test set accuracy is analyzed in previous works (Section \ref{sec:related_work}), this is the first comprehensive study of the influence of facial hair during training. First, effects of various amounts of facial hair in training set are measured. A controlled experiment is conducted by changing the facial hair percentage in a training set while maintaining the same set of identities. Then, data augmentation methods for manipulating pixels in the chin and mustache area are investigated.  The accuracy discrepancy between image pairs (clean-shaven against clean-shaven), (clean-shaven against facial hair) and (facial hair against facial hair) is evaluated. Results show that facial hair can significantly impact the recognition performance on African-Americans and Caucasians even when a perfectly balanced training set is utilized. Furthermore, this impact is observed to be different across races, potentially leading to a false judgement of demographic bias in facial recognition.

This paper is organized as follows. Section \ref{sec:related_work} reviews previous works on fairness in machine learning. Also, effect of facial hair presence during evaluation is discussed for face recognition. Section \ref{sec:real_training_data} presents the problem statement and methodology to address the accuracy bias caused by facial hair. Our dataset and evaluation protocol is also explained. Section \ref{sec:training_subset} describes our controlled training sets to measure the effect of facial hair during training. A data augmentation strategy on beard and mustache area is also investigated. Then, Section \ref{sec:casia} discusses the effect of the data augmentation procedure on a web-crawled dataset for a comparison with our controlled training set experiment. Finally, we conclude with a summary of the work and suggestions for future research.

\section{Related Work}
\label{sec:related_work}
Fairness is currently a topic of great interest in machine learning research. A comprehensive survey can be found in \cite{mehrabi2021survey}. Bias in learning algorithms can be broadly analyzed in two dimensions: algorithmic bias and data bias. In the first case, it is assumed that the training set is balanced, representing all subpopulations equally, and learning bias is considered to come from the learning procedure. Bias towards low complexity solutions in deep networks are investigated in recent works \cite{cao2021towards, rahaman2019spectral, zhang2021understanding, xu2019training, soudry2018implicit, poggio2018theory}.

While the lack of theoretical foundation in deep learning complicates the study of bias, data is often the root cause of biased decisions in machine learning \cite{kim2019learning, karkkainenfairface, wang2020towards, clark2019don, wu2023should}. Particularly, bias in face recognition models is extensively studied in many works \cite{robinson2023balancing, robinson2020face, wu2023face, drozdowski2020demographic, terhorst2021comprehensive, xu2021consistent, de2021fairness, neto2023compressed, yucer2022measuring}. Balancing the number of images in training across demographics is initially considered as a simple solution to the bias problem, but previous works \cite{wang2019balanced, albiero2020does} show that gender gap continues to exist even with a perfectly balanced training set. In addition to the effort on constructing a ``balanced'' training set, bias measurement strategies on the test set are also investigated in previous studies \cite{singh2022anatomizing, yucer2022measuring, cavazos2020accuracy, serna2021insidebias, karkkainenfairface}. 

While many of the aforementioned works investigate demographic biases across races, the effect of protected attributes obstruct the evaluation of fairness in face recognition \cite{terhorst2021comprehensive}. In \cite{wu2023face}, effects of illumination on African-American and Caucasian images are investigated. A pretrained BiSeNet \cite{yu2018bisenet} model is employed to segment a face image into 13 regions. Then, an image brightness metric is proposed to measure the face skin brightness. They show that comparison of two over-exposed (too bright) and comparison of two under-exposed (too dark) images can increase the match scores, resulting in higher FMR.

\textbf{Impact of facial hair on recognition bias.} 
Givens \etal \cite{givens2004features} analyze the effect of facial hair together with other attributes on non-deep learning face recognition approaches. 2,144 images from the FERET \cite{phillips1998feret} dataset are used in their experiments. A binary label is used to mark images as facial hair or clean-shaven. Their results suggest that when facial hair is present in one image and not the other, face recognition accuracy improves.
This can be interpreted as increased dissimilarity in an impostor pair of images reducing the similarity score, and therefore reducing the chance of a false match.

In \cite{lu2019experimental}, the impacts of seven covariates with facial hair are examined through the utilization of five Deep Neural Networks (DNNs). The assessment of facial hair is based on four binary labels: "no facial hair", "mustache", "goatee", and "beard". Their findings indicate that state-of-the-art (SOTA) deep learning models can accommodate variations in facial hair. However, an uncontrolled dataset is used in their evaluation.

Terhörst \etal \cite{terhorst2021comprehensive} presents a comprehensive analysis of facial attributes, including facial hair, with use of deep CNN face matchers.  In one particular result investigating the effect of 5 o'clock shadow on face images, they report better recognition performance with 5 o'clock shadow compared to clean-shaven images.

The examination of the impact of both scalp hair and facial hair on face recognition is analyzed in \cite{bhatta2023gender}. Scalp hair is detected by a segmentation model \cite{yu2018bisenet} and combination of Microsoft Face API and Amazon Rekognition predictions are used to classify images as clean-shaven and facial hair. They report that the accuracy of clean-shaven prediction is lower for African-Americans than for Caucasians, suggesting development of a better classifier for a more accurate analysis.
\begin{figure*}[htb]
        \begin{subfigure}[t]{.49\textwidth}
            \centering
            \includegraphics[width=.98\linewidth]{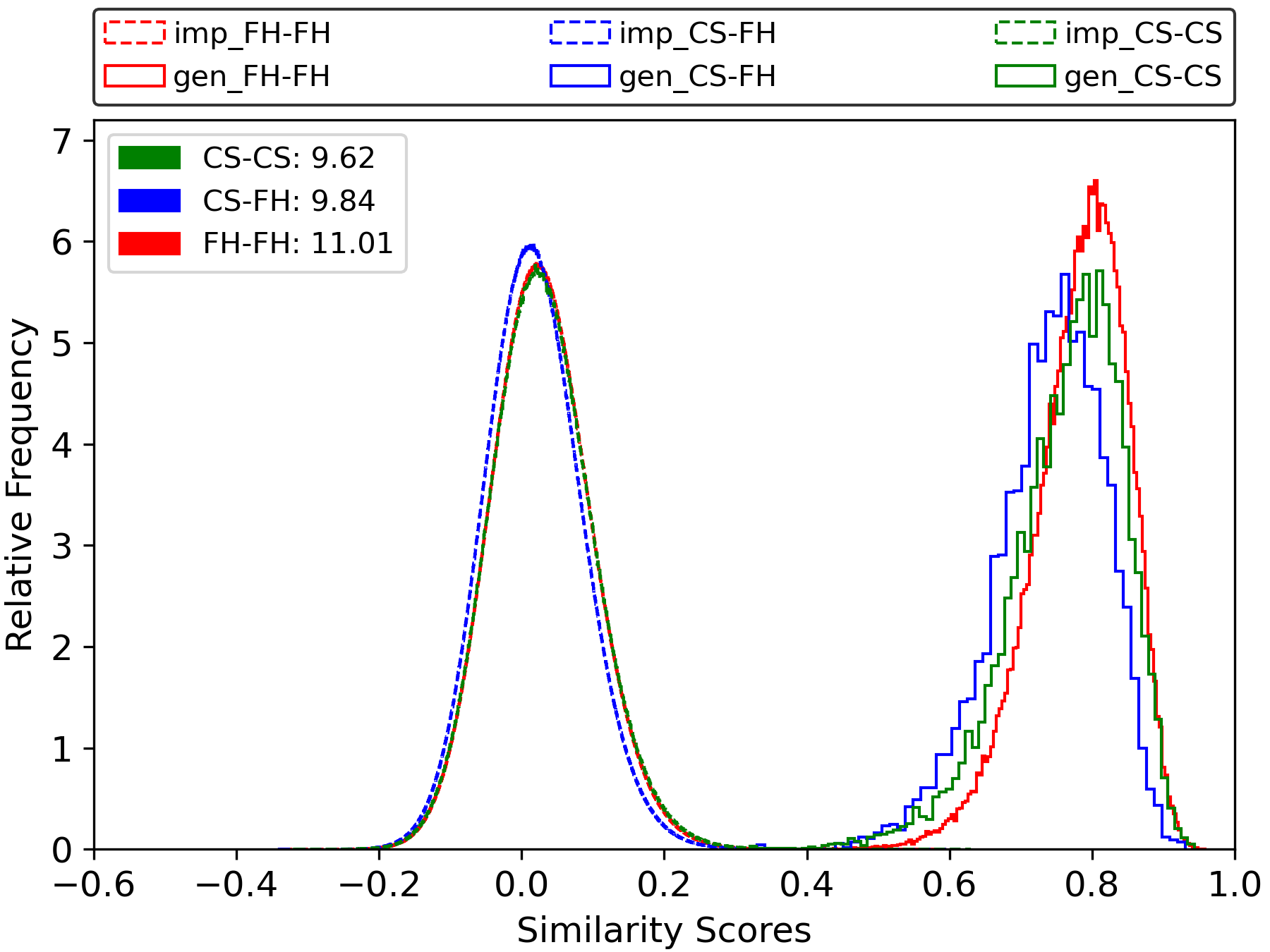}
            \caption{AAM}
        \end{subfigure}
        \begin{subfigure}[t]{.49\textwidth}
            \centering
            \includegraphics[width=.98\linewidth]{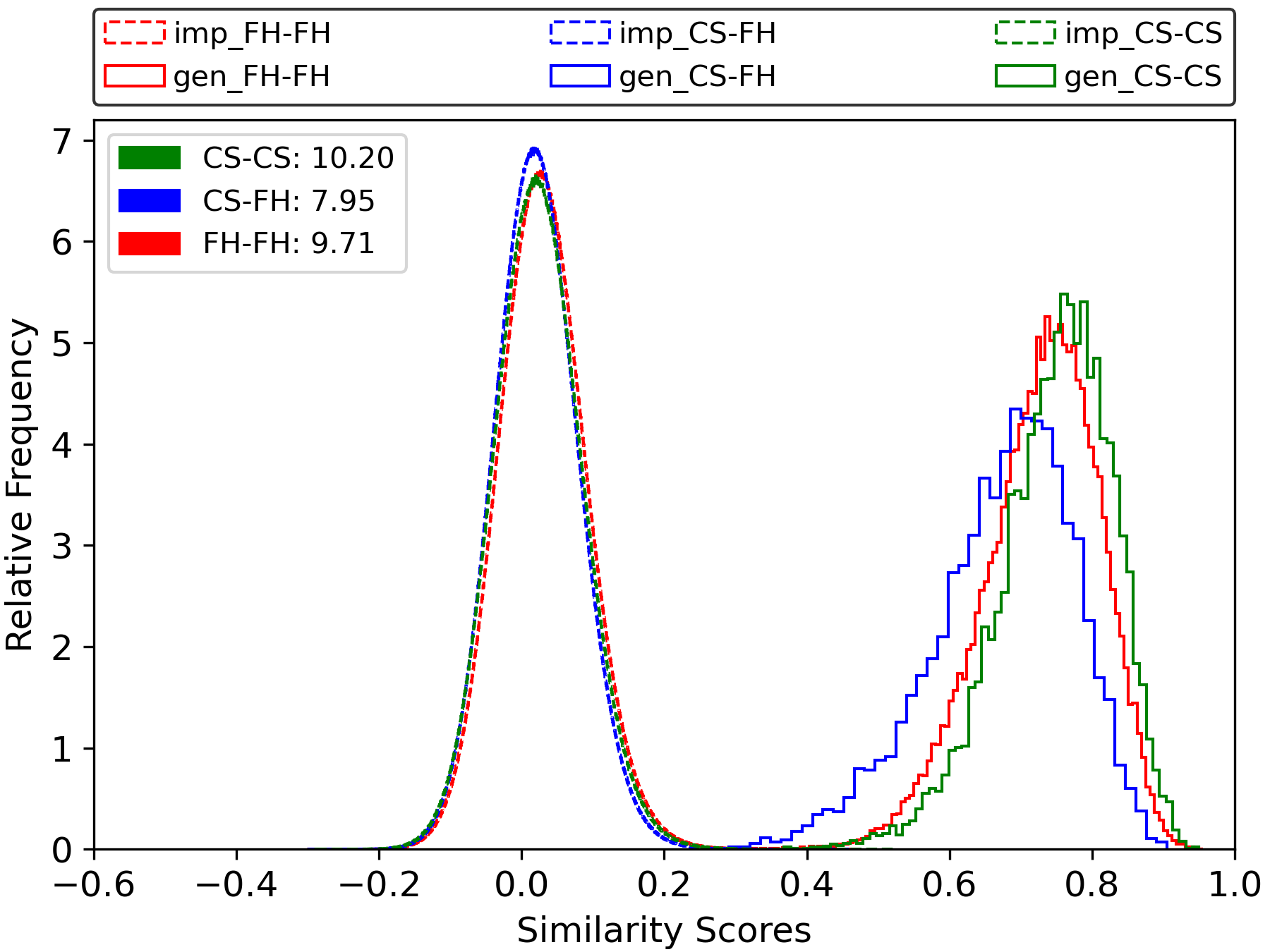}
            \caption{CM}

        \vspace{10pt} %
        \end{subfigure}        
        \begin{subfigure}[t]{.49\textwidth}
            \centering
            \includegraphics[width=.9\linewidth]{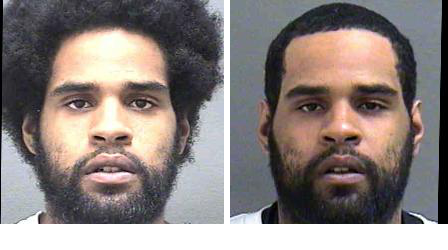}
            \vspace{-13pt}
            
            \caption*{similarity score: 0.81}
            
            \caption{Example AAM gen\_FH-FH}
        \end{subfigure}
        \hspace{0.01\textwidth}
        \begin{subfigure}[t]{.49\textwidth}
            \centering
            \includegraphics[width=.9\linewidth]{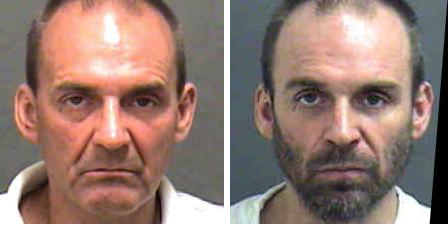}
            \vspace{-13pt}
            \caption*{similarity score: 0.44}
            \caption{Example CM gen\_CS-FH}
        \end{subfigure}

    \caption{Impostor and genuine distributions for CS-CS, CS-FH, and FH-FH image pairs. Similarity scores are obtained using a pretrained AdaFace model on WebFace12M. D-prime values are given at the upper-left of the plots. While the recognition performance is better for Caucasian males on CS-CS pairs (b), d-prime values are greater for African-American males on CS-FH and FH-FH pairs (a). Examples of AAM FH-FH (c) and CM CS-FH (d) genuine pairs are shown. }
    \label{fig:12m_scores}
\end{figure*}

In \cite{wu2023logical}, a facial hair dataset is presented and a facial hair attribute classifier is proposed. Facial hair information about the area and length can be obtained using a set of 22 binary labels. Using this classifier, the effect of beard region is analyzed \cite{wu2024facial}. They find mustache has a significant impact on recognition compared to facial hair on chin area, even though size of the mustache region is smaller.

A facial hair segmentation model is presented in \cite{ozturk2023beard}, trained using a hand-annotated facial hair dataset. Images are categorized according to the number of facial hair pixels. It is shown that a greater amount of facial hair can cause a significant recognition bias across demographics.

\section{Does recognition bias due to facial hairstyle decrease with larger training data?}
\label{sec:real_training_data}

We start our analysis by comparing the performance of face recognition models trained on different training sizes. Three publicly-available \cite{adaface_github} models, trained on Casia-WebFace \cite{yi2014learning}, WebFace4M and WebFace12M \cite{zhu2021webface260m} to minimize AdaFace loss \cite{kim2022adaface}, are used for obtaining face representations. To investigate the recognition bias between clean-shaven and facial hair images, a facial hair classifier is used \cite{wu2023logical}. The classifier can predict attributes about the area and length of facial hair. Beard area attributes: \emph{"no beard"}, \emph{"chin"}, \emph{"side to side"}. Mustache attributes: \emph{"no mustache"}, \emph{"mustache connected to beard"}, \emph{"mustache isolated"}. Length attributes: \emph{"5 o'clock shadow"}, \emph{"short"}, \emph{"medium"}, \emph{"long"}. We categorize images as clean-shaven (CS) and facial hair (FH) according to predictions of the model using a threshold value of $0.7$:
\[
    Images=
\begin{cases}
    CS, \textbf{ if }& \emph{"no beard"} \textbf{ and } \emph{"no mustache"}\\
    \\
    FH, \textbf{ if not} & \emph{("no beard"} \textbf{ or } \emph{"no mustache"}\\ & \textbf{ or } \emph{"5 o'clock shadow")}
\end{cases}
\]

\begin{figure}[htb]
    \centering
    \begin{subfigure}[t]{0.2\textwidth}
        \includegraphics[width=\linewidth]{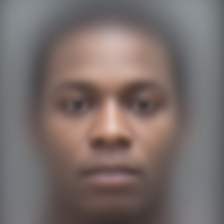}
        \caption{AAM CS}
    \end{subfigure}
        \centering
    \begin{subfigure}[t]{0.2\textwidth}
        \includegraphics[width=\linewidth]{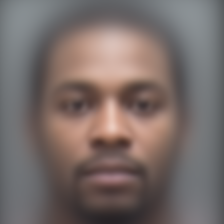}
        \caption{AAM FH}
        
    \end{subfigure}
        \begin{subfigure}[t]{0.2\textwidth}
        \includegraphics[width=\linewidth]{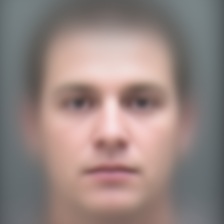}
        \caption{CM CS}
        
    \end{subfigure}
        \centering
    \begin{subfigure}[t]{0.2\textwidth}
        \includegraphics[width=\linewidth]{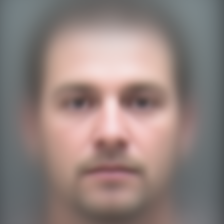}
        \caption{CM FH}
        
    \end{subfigure}
    
    \caption{Mean faces of clean-shaven (CS) and facial hair (FH) image sets for African-American (AAM) and Caucasian (CM) males on MORPH.}
    \label{fig:mean_faces}
\end{figure}

\begin{figure}[htb]
    \centering
    \includegraphics[width=.8\linewidth]{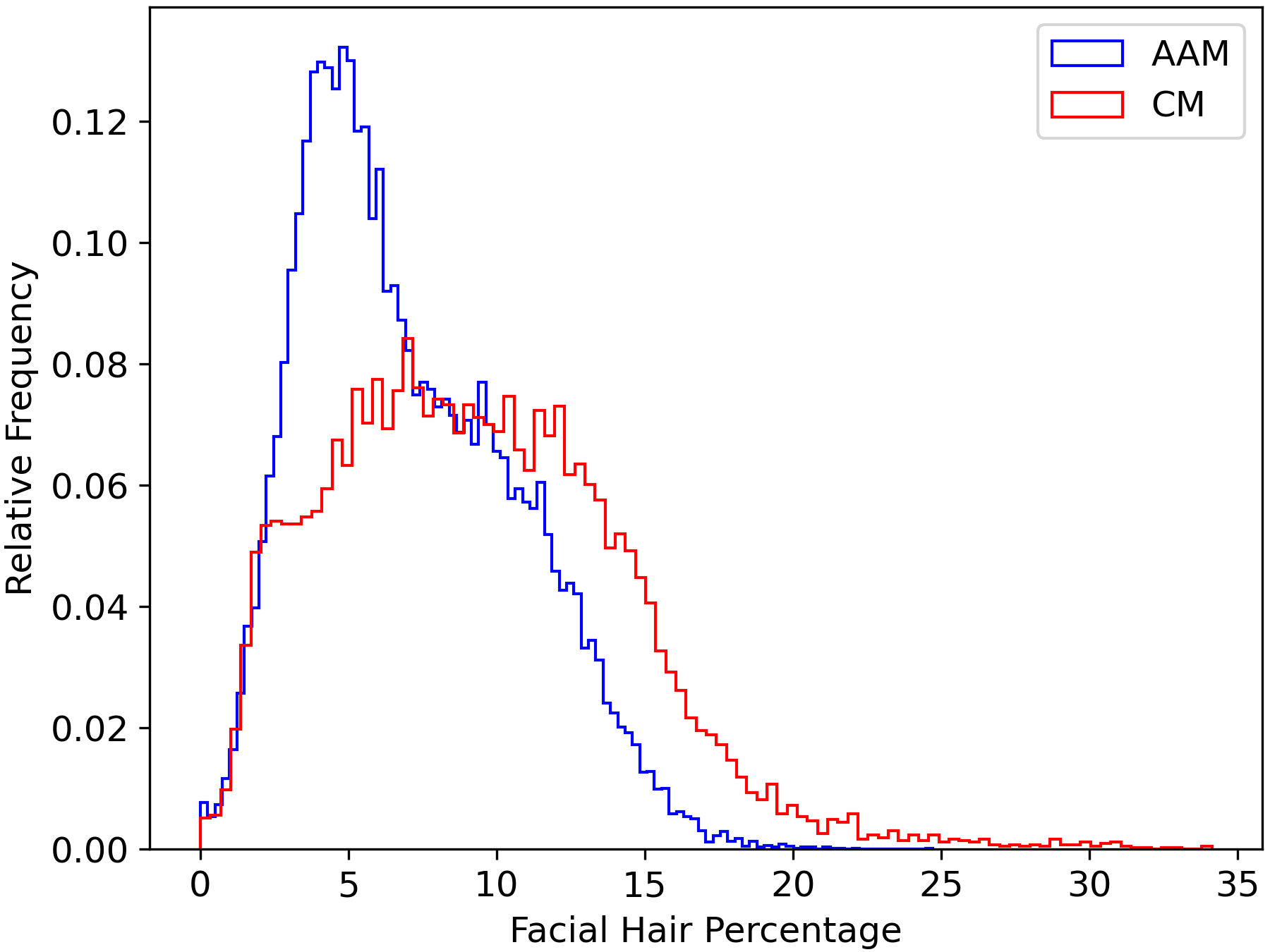}
    \caption{Percentage of number of facial hair pixels in a FH image for Caucasian and African-American males on MORPH.}
    \label{fig:fh_pixel_dist}
\end{figure}

\textbf{Dataset.} MORPH \cite{morph} is a widely used dataset to perform age, gender and race analyses \cite{wu2023face, wu2024facial, albiero2020does, bhatta2023gender}. MORPH images are captured in controlled conditions, as observed in passports, ID cards, etc., allowing a more detailed and accurate analysis of facial hair impact. The dataset consists of 56,245 images of 8,839 African-American males (AAM) and 35,276 images of 8,835 Caucasian males (CM). We use 5,123 CS and 24,968 FH images for AAM and 8,497 CS and 12,578 FH images CM to report results across demographics. Mean face images are shown in Figure \ref{fig:mean_faces}. The distribution of number of facial hair pixels, detected by a segmentation model \cite{ozturk2023beard} is given in Figure \ref{fig:fh_pixel_dist}. AAM have more images with a smaller fraction of facial hair pixels. Distributions may vary for other datasets.

\textbf{Evaluation.} First, image representations are obtained using a face recognition model. Then, cosine distance between images is measured to obtain similarity score distributions. We use d-prime to calculate the distance between genuine and impostor distributions. 
\expandafter\def\expandafter\normalsize\expandafter{%
    \normalsize%
    \setlength\abovedisplayskip{0pt}%
    \setlength\belowdisplayskip{8pt}%
    \setlength\abovedisplayshortskip{-8pt}%
    \setlength\belowdisplayshortskip{2pt}%
}
\newcommand{\meanA}{\bar{X}_1}
\newcommand{\meanB}{\bar{X}_2}
\newcommand{\stdA}{\sigma_1}
\newcommand{\stdB}{\sigma_2}
\newcommand{\cohenD}{
  \frac{\meanA - \meanB}{\sqrt{\frac{\stdA^2 + \stdB^2}{2}}}
}
\begin{equation}
    d' = \cohenD
    \label{eq:dprime}
\end{equation}

Higher d-prime values mean better separation of these distributions, indicating better recognition accuracy. Image pair group, CS-CS (clean-shaven against clean-shaven), CS-FH (clean-shaven against facial hair) and FH-FH (facial hair against facial hair) are created for evaluating performance discrepancy caused by facial hairstyle. We follow the same evaluation protocol for all the experiments throughout the paper.

Effects of training set size on the recognition performance for CS-CS, CS-FH and FH-FH is shown in Figure \ref{fig:training_size} across African-American (AAM) and Caucasian (CM) males. We evaluate the performance using 4 models, trained with 120K, 500K, 4M, 12M images. Even though d-prime values are increasing with larger training set size, there is a significant gap between particular groups (d-prime of AAM FH-FH is 11.01 while d-prime of CM CS-FH 7.95 using the model trained with 12M images) showing the impact of facial hairstyle in an image pair.

Impostor and genuine distributions for the AdaFace model trained on WebFace12M are given in Figure \ref{fig:12m_scores}. D-prime values are also given in the upper left. It can be seen that, when two images have different facial hairstyle (CS-FH), both impostor and genuine similarity scores decrease. However, the genuine distribution of CS-FH Caucasian males is significantly shifted to lower scores, resulting in the worst d-prime value among all pair groups. Note that, while d-prime of CS-CS  pairs is higher for Caucasian subjects, African-American subjects have better d-prime values for CS-FH and FH-FH image pairs. These results suggest that evaluation of recognition bias across races can be significantly influenced by the amount of clean-shaven and facial hair images in the test data.


\section{How does the facial hair distribution in training data affect the performance?}
\label{sec:training_subset}

To investigate the effect of facial hair distribution in training set, our goal is to have a set of male subjects that have both clean-shaven and facial hair images available to be used in the experiments. By doing so, we greatly control the other factors while changing the facial hair frequency in the training set; i.e., all training sets with different frequency of facial hair representation have the same set of identities and the same number of images per subject. The WebFace42M \cite{zhu2021webface260m} dataset is used to find such subjects. Clean-shaven and facial hair images are detected using the facial hair classifier \cite{wu2023logical} (see Section \ref{sec:real_training_data}).

\textbf{Training data.} Initially, we aim to have a training set size similar to Casia-WebFace \cite{yi2014learning} for comparison (10,575 subjects and 494,414 images). On WebFace42M, we found 5,000 subjects with 12 CS and 12 FH images (12-CS 12-FH). Subsets from this set of images are chosen to build our training sets. Facial hair frequency is controlled by selecting 12 images of these 5,000 male subjects. Additionally, we use the predictions of a gender classifier \cite{karkkainenfairface} to select 5,000 female subjects with 12 images to balance the number of images of male and female subjects. (Female subjects and images are chosen randomly from WebFace42M.)  The same 60,000 female images are used in each training set.

Training sets with different facial hair frequency are constructed by two approaches: (1) selecting \textbf{x} number of males with only CS images and \textbf{5,000-x} male subjects with only FH images. (2) selecting \textbf{x} CS images and \textbf{12-x} FH images for all 5,000 male subjects. In addition to the selected 60,000 male images, the same 60,000 images of 5,000 female subjects are included in the training set, resulting in 10,000 subjects and 120,000 images for all training sets.

\textbf{Implementation Details.} A ResNet-50 \cite{he2016deep} architecture is used to train models. Training lasts 30 epochs using 120,000 images with a batch size of 128 on a single GPU. Polynomial learning scheduler is employed with 3 warm-up epochs. Learning rate is set to 0.1. AdaFace \cite{kim2022adaface} loss is used for training. 

\subsection{Impact of facial hair variation across subjects}
\label{sec:across subjects}

For the first experiment, we created training subsets in which each male subject has either only clean-shaven or only facial hair images. We select \{0, 500, 1,000, 1,500, 2,000, 2,500, 3,000, 3,500, 4,000, 4,500, 5,000\} males with 12 CS images and \{5,000, 4,500, 4,000, 3,500, 3,000, 2,500, 2,000, 1,500, 1,000, 500, 0\} males with 12 FH images respectively to form 11 different sets. Training is repeated 5 times, selecting a random set of subjects, from our superset (12-CS and 12-FH), with only clean-shaven and subjects with only facial hair images. In the two cases, all images are clean-shaven ("5000-CS 0-FH") and all images are facial hair ("0-CS 5000-FH"), the same 120,000 images are used in all repetitions with random initialization.

\begin{figure*}[htb]
    \centering
    \begin{subfigure}[t]{\textwidth}
        \begin{subfigure}[t]{.48\textwidth}
            \includegraphics[width=\linewidth]{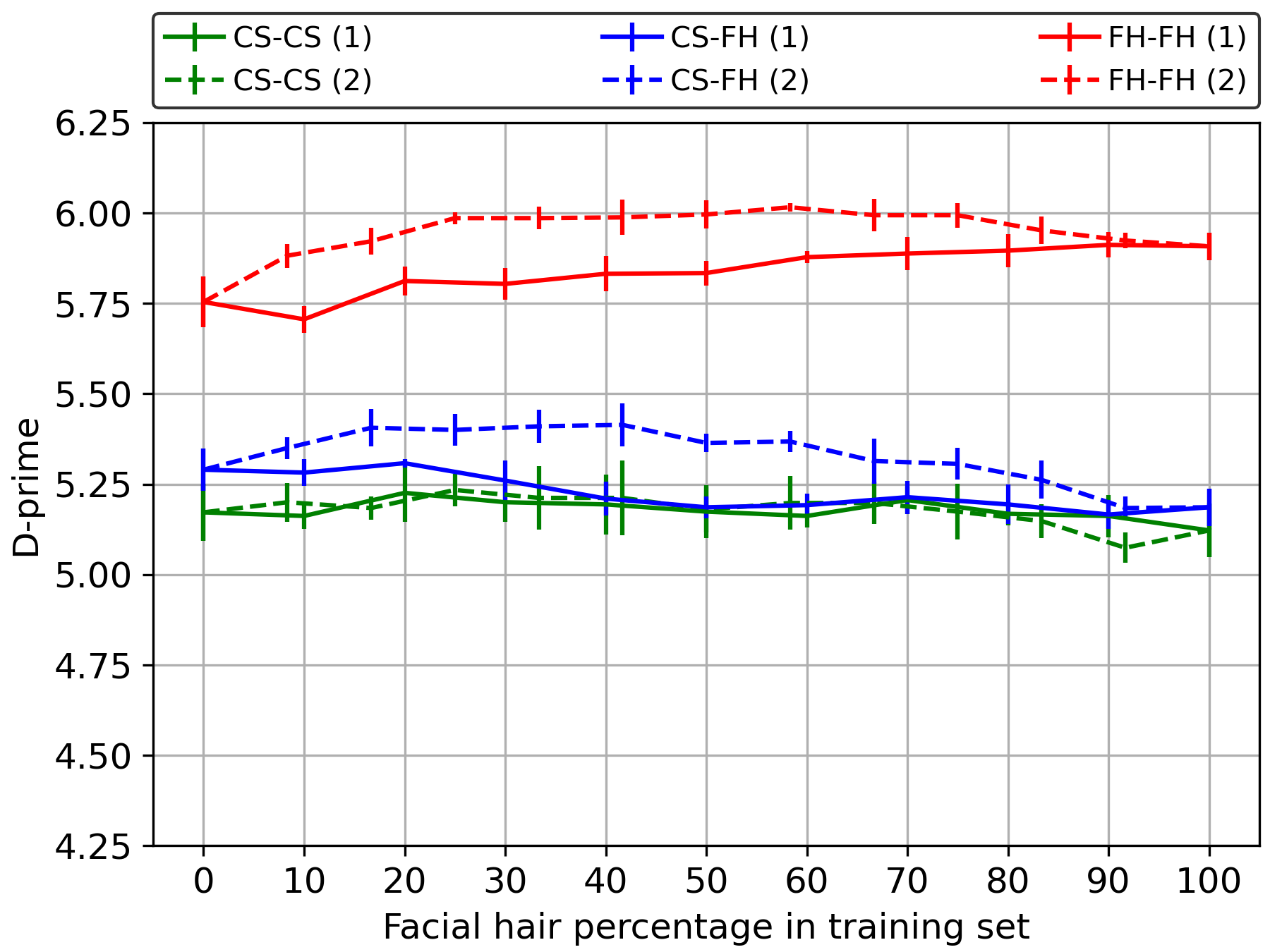}
            \caption{AAM}
        \end{subfigure}
            \begin{subfigure}[t]{.48\textwidth}
                \includegraphics[width=\linewidth]{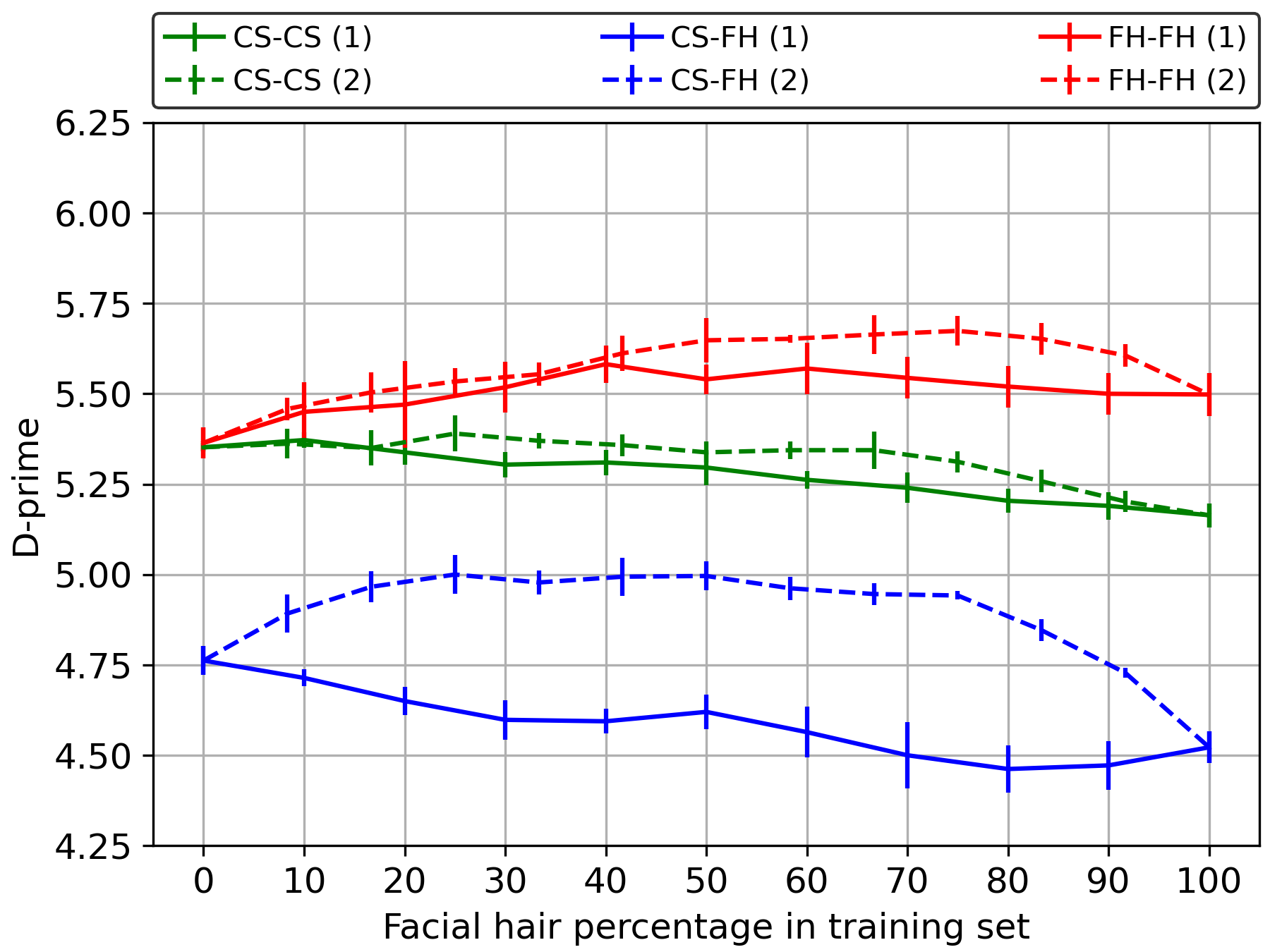}
                \caption{CM}
        \end{subfigure}
    \end{subfigure}
    \caption{Effect of facial hair distribution in training set. Recognition performance is measured on African-American (a) and Caucasian (b) males on MORPH. D-prime values are shown for CS-CS (clean-shaven versus clean-shaven), CS-FH (clean-shaven versus facial hair) and FH-FH (facial hair versus facial hair) image pairs. Dashed lines show the effect of facial hair ratio variation within subjects (Section \ref{sec:within subjects}) and solid lines shows the variation across subjects (Section \ref{sec:across subjects}). Vertical bars show the standard deviation of 5 repetition. Lower d-prime values are observed in most cases as facial hair percentage exceed $50\%$ in training data.}
    \label{fig:dprime_exp1_exp2}
\end{figure*}

\definecolor{mygreen}{rgb}{0.0, 0.5, 0.0}
\begin{table*}[htb]
\centering
\begin{tabular}{c|ccc|ccc}
\multirow{1}{*}{Train} & \multicolumn{6}{c}{Test} \\
\cline{2-7}
& \multicolumn{3}{c|}{AAM} & \multicolumn{3}{c}{CM} \\
\cline{2-7}
& CS-CS & CS-FH & FH-FH & CS-CS & CS-FH & FH-FH \\
\cline{1-7}
0-CS 12-FH & $5.12\pm0.07$ & $5.19\pm0.05$ & $5.91\pm0.04$ & $5.16\pm0.03$ & $4.52\pm0.04$ & $5.50\pm0.06$ \\
6-CS 6-FH & $5.18\pm0.06$ & $5.36\pm0.03$ &\textbf{$6.00\pm0.04$} & $5.34\pm0.03$ &\textbf{$5.00\pm0.04$} &\textbf{$5.65\pm0.06$} \\
12-CS 0-FH & $5.17\pm0.08$ & $5.29\pm0.06$ & $5.75\pm0.07$ & $5.35\pm0.01$ & $4.76\pm0.04$ & $5.36\pm0.04$ \\

\hline
\hline
\multicolumn{7}{l}{12-CS 0-FH w/ facial hair pixel aug. (male)} \\
\hline
prob. 0.2 & $5.19$ & $5.30$ & $5.85$ & $5.33$ & $4.79$ & $5.40$ \\
prob. 0.4 & $5.16$ & $5.37$ & $5.87$ & $5.28$ & $4.85$ & $5.45$ \\
prob. 0.6 & $5.08$ & $5.36$ & $5.83$ & $5.25$ & $4.93$ & $5.45$ \\
prob. 0.8 & $5.07$ & $5.30$ & $5.79$ & $5.16$ & $4.80$ & $5.36$ \\
prob. 1.0 & $5.13$ & $5.27$ & $5.82$ & $5.09$ & $4.73$ & $5.43$ \\

\hline
\hline
\multicolumn{7}{l}{12-CS 0-FH w/ facial hair pixel aug. (male and female)} \\
\hline
prob. 0.2 & $5.21$ & $5.39$ & $5.82$ & $5.34$ & $4.88$ & $5.41$ \\
prob. 0.4 & $5.10$ & $5.38$ & $5.84$ & $5.32$ & $4.90$ & $5.43$ \\
prob. 0.6 & $5.07$ & $5.34$ & $5.81$ & $5.27$ & $4.87$ & $5.39$ \\
prob. 0.8 & $5.12$ & $5.33$ & $5.80$ & $5.19$ & $4.86$ & $5.39$ \\
prob. 1.0 & $5.12$ & $5.35$ & $5.82$ & $5.23$ & $4.85$ & $5.44$ \\

\hline
\hline
\multicolumn{7}{l}{12-CS 0-FH w/ random pixel aug. (male)} \\
\hline
prob. 0.2 & $5.10$ & $5.32$ & $5.83$ & $5.37$ & $4.95$ & $5.49$ \\
prob. 0.4 & $5.19$ & $5.40$ & $5.90$ & $5.29$ & $4.84$ & $5.47$ \\
prob. 0.6 & $5.22$ & $5.33$ & $5.84$ & $5.22$ & $4.76$ & $5.36$ \\
prob. 0.8 & $5.20$ & $5.34$ & $5.85$ & $5.22$ & $4.81$ & $5.36$ \\
prob. 1.0 & $5.00$ & $5.13$ & $5.63$ & $5.13$ & $4.60$ & $5.20$ \\

\hline
\hline
\multicolumn{7}{l}{12-CS 0-FH w/ random pixel aug. (male and female)} \\
\hline
prob. 0.2 & $5.27$ & $5.34$ & $5.88$ & \cellcolor{mygreen!50}$5.46$ & $4.89$ &$5.41$ \\
\rowcolor{mygreen!50}
prob. 0.4 &$5.31$ &$5.43$ & $5.93$ & \cellcolor{mygreen!0}$5.38$ & $4.96$ & $5.51$ \\
prob. 0.6 & $5.07$ & $5.27$ & $5.77$ & $5.28$ & $4.83$ & $5.30$ \\
prob. 0.8 & $5.11$ & $5.28$ & $5.79$ & $5.24$ & $4.82$ & $5.37$ \\
prob. 1.0 & $5.07$ & $5.22$ & $5.76$ & $5.14$ & $4.73$ & $5.36$ \\

\end{tabular}
\caption{Effect of data augmentation on beard area. D-prime values between genuine and impostor distributions of image pairs (CS-CS, CS-FH and FH-FH) are given for African-American (AAM) and Caucasian (CM) males. The first three rows show the performance of models trained with: only facial hair images (0-CS 12-FH), balanced set (6-CS 6-FH) and only clean-shaven (12-CS 0-FH) images. Four data augmentation approaches are investigated on only clean-shaven set (12-CS 0-FH) with 5 different probability. Best performance gain is observed with random pixel augmentation to beard area on all male and female images.}
\label{tab:dprime_exp2}
\end{table*}

\subsection{Impact of facial hair variation within subjects}
\label{sec:within subjects}

In the second experiment, the frequency of facial hair is controlled by selecting \{0, 1, 2, 3, 4, 5, 6, 7, 8, 9, 10, 11, 12\} CS images and \{12, 11, 10, 9, 8, 7, 6, 5, 4, 3, 2, 1, 0\} FH images for each male subject to construct 13 different training subsets. Training is repeated 5 times, selecting a random set of images, from our superset (12-CS and 12-FH), for each subject. Note that, in the case of all images are CS ("12-CS 0-FH"), the training set is the same as with 5,000-CS 0-FH (Section \ref{sec:across subjects}). Similarly, "0-CS 12-FH" corresponds to "0-CS 5,000-FH".

\begin{figure*}[htb]
    \centering

    \begin{subfigure}[t]{.45\textwidth}
        \centering
        \begin{subfigure}[t]{.45\linewidth}
            \includegraphics[width=\linewidth]{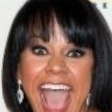}
        \end{subfigure}
        \hfill
        \begin{subfigure}[t]{.45\linewidth}
            \includegraphics[width=\linewidth]{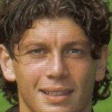}
        \end{subfigure}
            \caption{Clean-shaven images}
    \end{subfigure}
    \hfill
    \begin{subfigure}[t]{.45\textwidth}
        \centering
        \begin{subfigure}[t]{.45\linewidth}
            \includegraphics[width=\linewidth]{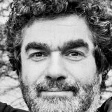}
        \end{subfigure}
        \hfill
        \begin{subfigure}[t]{.45\linewidth}
            \includegraphics[width=\linewidth]{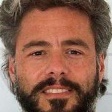}
        \end{subfigure}
            \caption{Facial hair images}
    \end{subfigure}

    \medskip

    \begin{subfigure}[t]{.45\textwidth}
        \centering
        \begin{subfigure}[t]{.45\linewidth}
            \includegraphics[width=\linewidth]{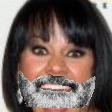}
        \end{subfigure}
        \hfill
        \begin{subfigure}[t]{.45\linewidth}
            \includegraphics[width=\linewidth]{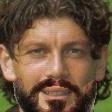}
        \end{subfigure}
            \caption{Augmentation using facial hair pixels}
    \end{subfigure}
    \hfill
    \begin{subfigure}[t]{.45\textwidth}
        \centering
        \begin{subfigure}[t]{.45\linewidth}
            \includegraphics[width=\linewidth]{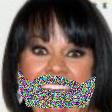}
        \end{subfigure}
        \hfill
        \begin{subfigure}[t]{.45\linewidth}
            \includegraphics[width=\linewidth]{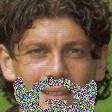}
        \end{subfigure}
            \caption{Augmentation using random pixels}
    \end{subfigure}
    
    \caption{Data augmentation using facial hair and random pixels. Clean-shaven images (a) are augmented using facial hair pixels (b) from Casia-WebFace. First, a facial hair mask in obtained using a segmentation model. Then the mask is warped to match facial landmarks between two images and facial hair pixels copied to a clean-shaven image (c). Instead of facial hair pixel values, random values are applied to the corresponding area in the second approach (d).}
    \label{fig:aug}
    
\end{figure*}

\textbf{Results.} Figure \ref{fig:dprime_exp1_exp2} shows the recognition performance for training subsets with different facial hair frequency. Dashed lines represent facial hair variation across subjects (Section \ref{sec:across subjects}), and solid lines are for facial hair variation within subjects (Section \ref{sec:within subjects}) in the training set. {\it It can be seen that, models trained with subjects that have both CS and FH images always perform better than the models trained with subjects that have only one type of facial hairstyle.} Overall, across 3 image pair groups and 2 races, d-prime values are higher when the facial hair frequency is  between $30\%$ and $70\%$. However, a performance gap still exits among CS-CS, CS-FH and FH-FH pairs, even when a balanced set (6-CS 6-FH) is used to train a face recognition model.  In this experiment, we do not find a frequency of facial hairstyle in the training data that makes a major change in the hairstyle-related accuracy difference in the test data. 

To further investigate the effects of facial hair representation in the training set, we explore two data augmentation approaches (Alg. \ref{alg:aug}). The first approach is copying facial hair pixels, detected by a facial hair segmentation model \cite{ozturk2023beard}, from the images in Casia-WebFace. We use 2,500 images that have more than $15\%$ of the pixels detected as facial hair. During training, a facial hair image is picked randomly and facial hair pixels are copied to a clean-shaven image. Facial landmark points \cite{bulat2017far} are used to warp the facial hair mask to match points between two images. 
\begin{algorithm}
    \KwData{CS\_image, FH\_image}
    \KwResult{Augmented\_image}

    \BlankLine
    FH\_mask $\leftarrow$ segmentFacialHair(FH\_image)\;

    CS\_landmarks $\leftarrow$ predictLandmarks(CS\_image)\;
    FH\_landmarks $\leftarrow$ predictLandmarks(FH\_image)\;

    Warped\_FH\_mask $\leftarrow$ warpImage(FH\_mask, CS\_landmarks, FH\_landmarks)\;

    Augmented\_image $\leftarrow$ replacePixels(CS\_image, Warped\_FH\_mask)\;

    \Return Augmented\_image\;
    
    \caption{Augment clean-shaven image}
    \label{alg:aug}
\end{algorithm}
In the second approach, instead of transferring real facial hair pixels from a different image, we apply random pixel values to a target image \cite{zhong2020random}. Examples are shown in Figure \ref{fig:aug}.

The impact of data augmentation on recognition performance for CS-CS, CS-FH and FH-FH pair groups is reported in Table \ref{tab:dprime_exp2}. Augmentation is applied to every image in the clean-shaven training set (12-CS 0-FH) with the probability of 0.2, 0.4, 0.6, 0.8 and 1.0. In addition, results of augmentation on only male images against augmenting all images (male and female) is compared. We find that, setting random pixel values in the beard area can be useful to increase recognition accuracy not only on facial hair images but also on CS-CS pairs. It can be seen that d-prime values are increasing over the baseline (12-CS 0-FH) and even exceed the performance of perfectly balanced set (6-CS 6-FH) for AAM CS-CS, AAM CS-FH and CM CS-CS image pairs.

\section{Facial hair area and Casia-WebFace}

\definecolor{mygreen}{rgb}{0.0, 0.5, 0.0}
\begin{table*}[h]
\centering
\begin{tabular}{c|ccc|ccc}
\multirow{1}{*}{Train} & \multicolumn{6}{c}{Test} \\
\cline{2-7}
& \multicolumn{3}{c|}{AAM} & \multicolumn{3}{c}{CM} \\
\cline{2-7}
& CS-CS & CS-FH & FH-FH & CS-CS & CS-FH & FH-FH \\
\cline{1-7}
Casia-WebFace wo/ aug. & $6.20\pm0.05$ & $6.29\pm0.02$ & $7.10\pm0.02$ & $6.86\pm0.02$ & $6.11\pm0.09$ & $6.82\pm0.05$ \\

\hline
\hline
\multicolumn{7}{l}{Casia-WebFace w/ random pixel aug. instead of facial hair  (male)} \\
\hline
prob. 0.2 & $6.20$ & $6.38$ & $7.12$ &\cellcolor{mygreen!50}$6.94$ & $6.10$ & $6.81$ \\
prob. 0.4 & $6.24$ & $6.35$ & $7.10$ & $6.90$ & $6.09$ & $6.77$ \\
prob. 0.6 & $6.17$ & $6.28$ & $7.04$ & $6.90$ & $6.11$ & $6.88$ \\
prob. 0.8 & $6.22$ & $6.30$ & $7.08$ & $6.91$ & $6.17$ & $6.83$ \\
prob. 1.0 & $6.19$ & $6.32$ & $7.16$ & $6.88$ & $6.06$ & $6.82$ \\

\hline
\hline
\multicolumn{7}{l}{Casia-WebFace w/ random pixel aug. mustache area (male)} \\
\hline
prob. 0.2 & $6.20$ & $6.26$ & $7.14$ & $6.87$ & $6.11$ & $6.89$ \\
prob. 0.4 & $6.29$ & $6.35$ & $7.11$ & $6.84$ & $6.08$ & $6.81$ \\
prob. 0.6 & $6.17$ & $6.21$ & $7.04$ & $6.75$ & $6.13$ & $6.77$ \\
prob. 0.8 & $6.08$ & $6.29$ & $7.05$ & $6.70$ & $6.04$ & $6.79$ \\
prob. 1.0 & $5.87$ & $5.80$ & $6.44$ & $6.30$ & $5.60$ & $6.19$ \\

\hline
\hline
\multicolumn{7}{l}{Casia-WebFace w/ random pixel aug. mustache area (male and female)} \\
\hline
prob. 0.2 & $6.28$ & $6.37$ & $7.15$ & $6.81$ & $6.14$ & $6.89$ \\
prob. 0.4 & $6.24$ & $6.28$ & $7.13$ & $6.80$ & $6.13$ & $6.86$ \\
prob. 0.6 & $6.16$ & $6.30$ & $7.10$ & $6.72$ & $6.00$ & $6.78$ \\
prob. 0.8 & $6.12$ & $6.29$ & $7.04$ & $6.58$ & $5.99$ & $6.76$ \\
prob. 1.0 & $5.65$ & $5.86$ & $6.60$ & $6.24$ & $5.66$ & $6.24$ \\

\hline
\hline
\multicolumn{7}{l}{Casia-WebFace w/ random pixel aug. mustache and beard area (male)} \\
\hline
prob. 0.2 & $6.23$ & $6.34$ & $7.13$ & $6.83$ &\cellcolor{mygreen!50}$6.24$ &\cellcolor{mygreen!50}$6.93$ \\
prob. 0.4 & $6.16$ & $6.27$ & $7.05$ & $6.81$ & $6.17$ & $6.80$ \\
prob. 0.6 & $6.15$ & $6.39$ & $7.09$ & $6.71$ & $6.02$ & $6.66$ \\
prob. 0.8 & $6.02$ & $6.26$ & $6.95$ & $6.62$ & $6.05$ & $6.67$ \\
prob. 1.0 & $5.54$ & $5.54$ & $6.05$ & $5.99$ & $5.14$ & $5.57$ \\

\hline
\hline
\multicolumn{7}{l}{Casia-WebFace w/ random pixel aug. mustache and beard area (male and female)} \\
\hline
prob. 0.2 &\cellcolor{mygreen!50}$6.29$ &\cellcolor{mygreen!50}$6.45$ &\cellcolor{mygreen!50}$7.19$ & $6.85$ & $6.20$ & $6.82$ \\
prob. 0.4 & $6.15$ & $6.34$ & $7.07$ & $6.76$ & $6.16$ & $6.87$ \\
prob. 0.6 & $6.04$ & $6.26$ & $6.97$ & $6.71$ & $6.07$ & $6.65$ \\
prob. 0.8 & $6.05$ & $6.26$ & $6.88$ & $6.52$ & $5.95$ & $6.55$ \\
prob. 1.0 & $5.41$ & $5.60$ & $6.27$ & $5.86$ & $5.26$ & $5.95$ \\

\end{tabular}
\caption{Effect of data augmentation on beard and mustache area during training is analyzed on Casia-WebFace. Augmentation results are reported for 5 range of probability. Comparison between augmentation on only male against all images is given.}
\label{tab:dprime_exp1}
\end{table*}

\label{sec:casia}

We further investigate the effect of beard area manipulation on Casia-WebFace to measure the effect on a larger training dataset. First, a face recognition model is trained without data augmentation as baseline. Training is repeated 5 times with random initiation. Mean and standard deviation is reported in Table \ref{tab:dprime_exp1}. We first conduct an experiment to investigate usage of random pixel values on a beard area, instead of real facial hair pixels. A facial hair segmentation model \cite{ozturk2023beard} is used to segment facial hair region on 490,623 images. Facial hair masks for 93,273 images are obtained. Pixels labeled as facial hair are replaced with random pixel values with the 5 levels of probability (0.2, 0.4, 0.6, 0.8, 1.0) for each image during training. It can be seen in Table \ref{tab:dprime_exp1} (\emph{Casia-WebFace w/ random pixel aug. instead of facial hair (male)}) that even replacing all real facial hair pixels with random values does not have a significant effect over the baseline (\emph{Casia-WebFace wo/ aug.}). The best results for CM CS-CS image pairs (6.94) is observed with this data augmentation approach with the probability of 0.2. 

Next, we apply random pixel augmentation procedure to only mustache area and both beard and mustache area to measure the importance of the particular face area to be augmented. Comparison between data augmentation on only male images against augmenting all male and females images is also reported. 68 facial landmark points \cite{bulat2017far} are used to define mustache and beard area for each image. Lower performance is observed for all cases as the probability of applying augmentation increases. The best results are achieved augmenting mustache and beard area on all images with probability of 0.2.

\section{Conclusion and Discussion}

We investigate the recognition bias (accuracy difference) that arises due to clean-shaven and facial hair face images. First, we show that even though larger training sets increase the performance for all image pairs, the accuracy gap caused by facial hairstyles does not decrease; see Figure \ref{fig:training_size}. Going from 4M to 12M training images does not significantly improve results for CM CS-FH. To investigate this discrepancy, training sets with varying facial hair amount are constructed. Results show that the accuracy gap does not diminish even with a perfectly balanced set.

Our findings on the effect of facial hair show that balanced training set construction either using only real images or using a data augmentation method that targets only the facial hair area on face images does not significantly mitigate bias caused by facial hairstyle. Note that the accuracy discrepancy is observed on African-American and Caucasian images in different amounts. While better recognition performance is measured on CS-CS image pairs for Caucasian male subjects, genuine similarity scores of CS-FH and FH-FH pairs are higher for African-American resulting in better accuracy. This suggests that judgement on whether recognition bias exists between demographics is heavily influenced by the facial hair present in the evaluation set and rigorous methodologies are necessary for bias assessment.

{\small
\bibliographystyle{ieee_fullname}
\bibliography{egbib}
}

\end{document}